\def\year{2020}\relax
\documentclass[letterpaper]{article}
\usepackage{aaai20}
\usepackage{times}
\usepackage{helvet}
\usepackage{courier}
\usepackage{todonotes}
\usepackage{epstopdf}
\usepackage{hyperref}

\usepackage{algorithm, algpseudocode}

\newcommand{\idest}{{\it i.e.}}
\newcommand{\exemp}{{\it e.g.}}
\newcommand{\etc}{{\it etc.}}
\newcommand{\etal}{{\it et al.}}

\begin{document}

\title{HAPRec: Hybrid Activity and Plan Recognizer}
\author{Roger Granada \and Ramon Fraga Pereira \and Juarez Monteiro \and Leonardo Amado\\ {\bf Rodrigo Barros \and Duncan Ruiz \and Felipe Meneguzzi}\\
School of Informatics - Pontifical Catholic University of Rio Grande do Sul \\
Porto Alegre - RS, Brazil\\
\{roger.granada, ramon.pereira, juarez.santos, leonardo.amado\}@acad.pucrs.br\\
\{rodrigo.barros, duncan.ruiz, felipe.meneguzzi\}@pucrs.br\\}

\maketitle

\begin{abstract}
Computer-based assistants have recently attracted much interest due to its applicability to ambient assisted living. 
Such assistants have to detect and recognize the high-level activities and goals performed by the assisted human beings. 
In this work, we demonstrate activity recognition in an indoor environment in order to identify the goal towards which the subject of the video is pursuing. 
Our hybrid approach combines an action recognition module and a goal recognition algorithm to identify the ultimate goal of the subject in the video. 
\end{abstract}

\section{Introduction}

Activity recognition can be understood as the task of recognizing the independent set of actions that generates an interpretation of a movement that is being performed. 
On the other hand, plan recognition can be understood as the task of recognizing agent goals and plans based on observed interactions in an environment. 
These observed interactions can be either events provided by sensors or actions/activities performed by an agent. 
Although much research effort focuses on activity and plan recognition as separate challenges, comparatively less effort focused on attempting to identify higher-level plans from activities in video sequences, \idest, try to understand the overarching goal of subjects within a video and make the correct inference from the observed activities. 
Rafferty \etal~\shortcite{RaffertyEtAl2017} use sensor based approach to implement assistive smart homes. 
Their approach is based upon an intention recognition mechanism that uses sensors affixed to objects and an ontological rule-based goal recognition system. 
Massardi \etal~\shortcite{MassardiEtAl2019} performs plan recognition using plan libraries from learned activities. Their top-down approach uses a particle filter with a population of plan trees to deal with noisy observations, producing a quick reliable solution.

In this work, we develop a hybrid approach that comprises both activity and plan recognition that identifies, from a set of candidate plans, which plan a human subject is pursuing based exclusively on still-camera video sequences.
To recognize such plan, we employ an activity recognition algorithm based on convolutional neural networks (CNN), which generates a sequence of activities that are checked for temporal consistency against a plan library using a symbolic plan recognition approach modified to work with a CNN. 
As supplemental material we provide a video demonstration of our architecture\footnote{Link to our video: \url{https://youtu.be/eb_6I6dzrEE}}.

\begin{figure}[t!]
    \centering
    \includegraphics[width=0.5\textwidth]{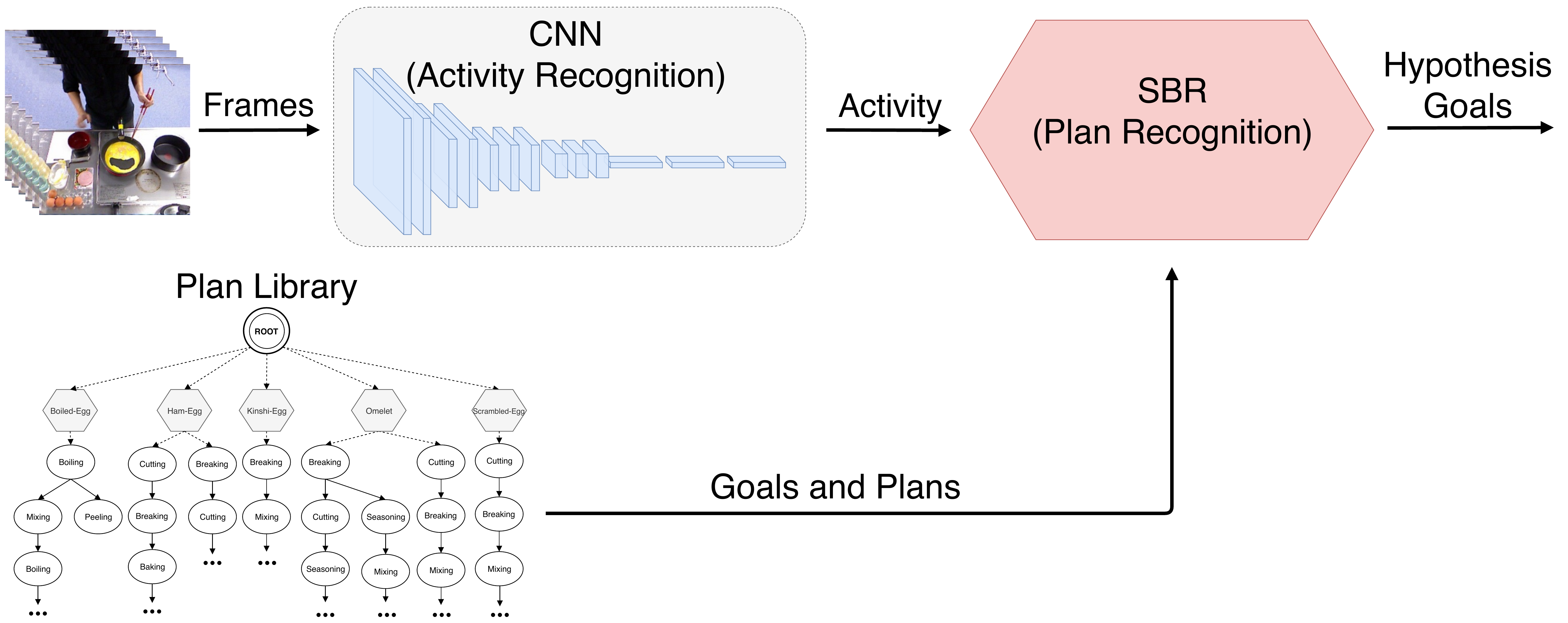}
    \caption{Pipeline of the hybrid architecture for activity and plan recognition.}
    \label{fig:pipeline}
\end{figure}

\section{A Hybrid Architecture for Activity and Plan Recognition}
\label{sec:method}

Our hybrid architecture is divided in two main parts: i) CNN-based activity recognition, and ii) CNN-backed symbolic plan recognition. 
The first part consists of training a Convolutional Neural Network (CNN) using video frames as input, with the activity being done in the video as the expected output. 
Our CNN architecture is based on GoogLeNet architecture \cite{SzegedyEtAl2015} and computes a probability score for all possible classes (\emph{softmax} output) for each frame. 
If two classes contain a high probability and the difference between them is lower than a threshold ($\theta$), we use an heuristic approach to disambiguate classes. 
This heuristic consists of assigning the class of the last frame to the current frame in case one of the two classes is equal to the class of the last frame. 
Otherwise, the current frame receives the class that contains the highest probability, disregarding the threshold.

After using the CNN to identify the activity being pursued, we use a plan recognizer that returns a set of possible plans that are temporally consistent with what is recognized from the input frames. 
To perform the task of plan recognition, we use a symbolic plan recognition approach called Symbolic Behavior Recognition (SBR). 
SBR \cite{ZilberbrandKaminka2005} is a plan recognition approach that takes as input a plan library and a sequence of observations, in this case, a sequence of observed feature values. 
Feature values are used as a set of conditions to execute a plan-step in a plan library. 
To match observed features with plan-steps in a plan library, we use an efficient matching step that maps observed features with matching plan-step nodes in a plan library. 
To do so, they use a feature decision tree (FDT) that maps observable features to plan-steps in a plan library. 
As output, SBR returns set of hypotheses plans such that each hypothesis represents a plan that achieves a top-level goal in a plan library. 
Instead of using the FDT to match observations with consistent plan-steps in the plan library, we modify the SBR and replace the FDT with the CNN-backed Activity Recognition. 
For instance, given a video frame, the CNN-based Activity Recognition returns which activity such video frame corresponds, and subsequently, we take this activity as input to the SBR, as shown in Figure~\ref{fig:pipeline}. 
Note that to recognize goals and plans using the SBR, we must model a plan library containing a set of possible sequence of activities (\idest, plan) that achieves goals. 
In this paper, a plan library corresponds to a model that contains a set of plans to achieve cooking menus.

\section{Application}
\label{sec:application}

\begin{figure}[t!]
    \centering
    \includegraphics[width=0.44\textwidth]{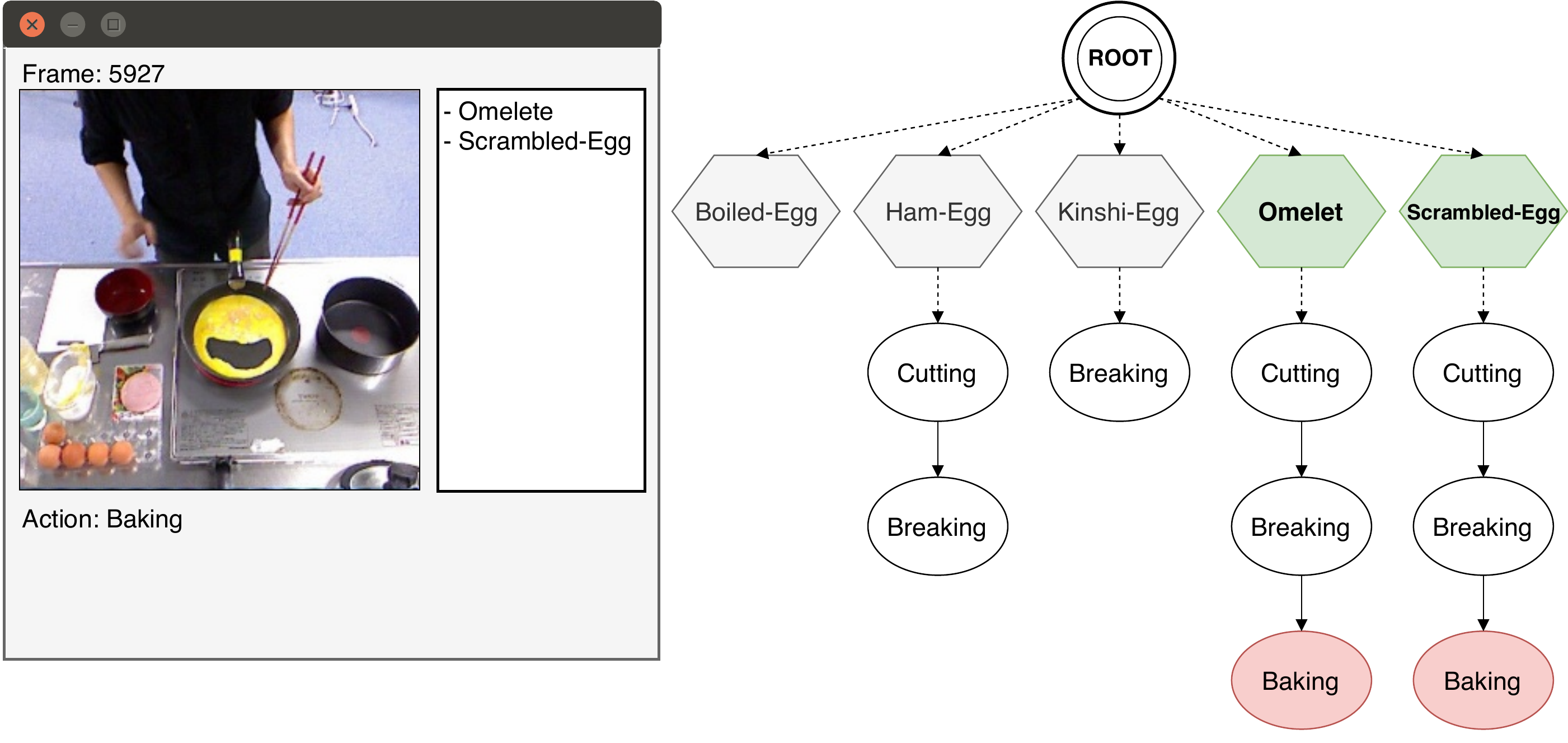}
    \caption{Demo screen showing the activity identified in the current frame, the plan-steps and the set of candidate goals.}
    \label{fig:demo}
\end{figure}

We create \emph{HAPRec}~\cite{Granada2017_PAIR} for demonstrating that it is possible to perform goal recognition using CNN-based activity recognition and plan libraries with real-world data (images). 
To demonstrate our work, we use the activities from ICPR 2012 Kitchen Scene Context based Gesture Recognition dataset (KSCGR) \cite{ShimadaEtAl2013}, which contains video sequences of five menus for cooking eggs in Japan: \emph{Ham and Eggs}, \emph{Omelet}, \emph{Scrambled Egg}, \emph{Boiled Egg}, and \emph{Kinshi-Tamago}. 
Each menu is performed by 7 subjects: 5 actors in training datasets and 2 actors in evaluation datasets, \idest, 5 cooking scenes are available for each training menu. 
Eight cooking gestures composes the dataset: \emph{breaking}, \emph{mixing}, \emph{baking}, \emph{turning}, \emph{cutting}, \emph{boiling}, \emph{seasoning}, \emph{peeling}, and \emph{none}, where \emph{none} means that there is not an activity being performed in the current frame. 
We chose the KSCGR dataset since it contains the activity being performed in each frame (\exemp, \emph{breaking}, \emph{baking} and \emph{turning}) as well as the goal achieved in the whole video sequence (\exemp, preparing \emph{Ham and Eggs}, \emph{Omelet}, \emph{Scrambled Egg}, \etc). 
Thus, we can carry out activity recognition using activities performed in each frame and plan recognition using the steps to achieve the recipe in each video. 
For recognizing goals and plans, we model a plan library containing knowledge of the agent's possible goals and plans based on the dataset, where each recipe is a top-level goal in the plan library. 
Based on videos from the training set, we model all possible plans for achieving each possible menu (\idest, top-level goal). 
We consider that a sequence of cooking gestures is analogous to a sequence of plan-steps, \idest, a plan in the plan library. 
Figure~\ref{fig:demo} illustrates the demo screen, showing the current image of the dataset, its frame id, the action predicted in that frame (\emph{Baking}), and the list of candidate goals (\emph{Omelet} and \emph{Scrambled-Egg}). 
On the right side, the sequence of plan-steps identified and the top-level goals, where the candidate goals are highlighted in green. 

\section{Conclusion}
\label{sec:conclusion}

We presented \emph{HAPRec}, a tool that performs both activity and plan recognition using real-world data. 
Our architecture is based on CNNs and a modified symbolic approach to plan recognition. 
We demonstrated how the algorithm works by testing using a kitchen scene environment containing actions performed by subjects and plans (recipes).

\section*{Acknowledgement}
This study was financed in part by Coordena\c{c}\~ao de Aperfei\c{c}oamento de Pessoal de N\'ivel Superior (CAPES) and the CAPES/FAPERGS agreement (DOCFIX 04/2018). 
We gratefully acknowledge the support of NVIDIA Corporation with the donation of the Titan Xp GPU.

\bibliography{references}
\bibliographystyle{aaai}

\end{document}